\documentclass[sigconf,nonacm]{acmart}
\AtBeginDocument{%
  }

\setcopyright{acmlicensed}
\copyrightyear{2024}
\acmYear{2024}
\acmDOI{XXXXXXX.XXXXXXX}

\acmISBN{978-1-4503-XXXX-X/18/06}

\usepackage{url}            % simple URL typesetting
\usepackage{booktabs}       % professional-quality tables
\usepackage{amsfonts}       % blackboard math symbols
\usepackage{nicefrac}       % compact symbols for 1/2, etc.
\usepackage{microtype}      % microtypography
\usepackage{xcolor}         % colors
\usepackage{makecell}
\usepackage{graphicx}
\usepackage{multirow}
\usepackage{multicol}
\usepackage{chngcntr}
\usepackage{subcaption}
\usepackage[bottom]{footmisc}
\usepackage{lipsum} 
\usepackage{wrapfig}
\usepackage[utf8]{inputenc}
\usepackage{algorithm}
\usepackage{algpseudocode}
\usepackage{setspace}

\begin{document}

\title{Optimizing Electric Vehicle Charging Station Locations: A Data-driven System with Multi-source Fusion}


\author{Lihuan Li*}
\affiliation{%
  \institution{UNSW}
  \city{Sydney}
  \country{Australia}}
\email{lihuan.li@student.unsw.edu.au}

\author{Du Yin*}
\affiliation{%
  \institution{UNSW}
  \city{Sydney}
  \country{Australia}}
\email{du.yin@unsw.edu.au}

\author{Hao Xue}
\affiliation{%
  \institution{UNSW}
  \city{Sydney}
  \country{Australia}}
\email{hao.xue1@unsw.edu.au}

\author{David Lillo-Trynes}
\affiliation{%
  \institution{Compass IoT}
  \city{Sydney}
  \country{Australia}}
\email{David.Lt@Compassiot.com.au}

\author{Flora Salim}
\affiliation{%
  \institution{UNSW}
  \city{Sydney}
  \country{Australia}}
\email{flora.salim@unsw.edu.au}

\thanks{* Equal contribution}
\renewcommand{\shortauthors}{Li et al.}

\begin{abstract}
With the growing electric vehicles (EVs) charging demand, urban planners face the challenges of providing charging infrastructure at optimal locations. 
For example, range anxiety during long-distance travel and the inadequate distribution of residential charging stations are the major issues many cities face. 
To achieve reasonable estimation and deployment of the charging demand, we develop a data-driven system based on existing EV trips in New South Wales (NSW) state, Australia, incorporating multiple factors that enhance the geographical feasibility of recommended charging stations. 
Our system integrates data sources including EV trip data, geographical data such as route data and Local Government Area (LGA) boundaries, as well as features like fire and flood risks, and Points of Interest (POIs). 
We visualize our results to intuitively demonstrate the findings from our data-driven, multi-source fusion system, and evaluate them through case studies. 
The outcome of this work can provide a platform for discussion to develop new insights that could be used to give guidance on where to position future EV charging stations.
% The outcome of this work could provide valuable insights and guide decisions on the location of EV charging stations.

\end{abstract}


% \ccsdesc[500]{Human-centered computing~Ubiquitous and mobile computing}



\maketitle

\section{Introduction}
With the increasing greenhouse gas emissions and air pollution, electric vehicles (EVs) using renewable energy are considered a potential substitution for gasoline-powered vehicles to ensure sustainable development and mitigate climate change. According to the New South Wales (NSW) government~\footnote{\url{https://www.energy.nsw.gov.au/households/rebates-grants-and-schemes/vehicle-emissions-offset-scheme-veos/your-quick-guide}} in Australia, current gasoline-powered light vehicles in New South Wales (NSW) produce around 65\% of transport emissions. As reported by Australian Electric Vehicle Council~\footnote{\url{https://electricvehiclecouncil.com.au/wp-content/uploads/2022/10/State-of-EVs-October-2022.pdf}}, the EV market share has increased to 8.5\% in 2023, which is 123.7\% higher than 2022. EV sales in NSW reached 9.0\%, increasing by 130.7\% from 2022, and are still expected to grow rapidly. Therefore, strategic placement of charging stations becomes essential to cope with the large charging demand.

% \textcolor{red}{1. data-driven, 2. accessibility (fire, flood, POIs, Routes), 3. LGA,.}

Previous studies~\cite{ahn2015analytical,lu2021data} tackle the EV charging station recommendation problem by building objective functions to minimize the EV driver charging cost and the charging station deployment cost. Nowadays, the proliferation of GPS trajectory data collected reflects daily travel patterns, further evidently illustrating the regional charging demand. Some works are based on taxi~\cite{hehuan2021location,liu2019data} and private vehicle~\cite{dong2014charging} trajectories to generate charging stations by simulating future charging demand. However, these data-driven methods neglect a more comprehensive evaluation of the accessibility of recommended charging stations. First, a charging station should be accessible, reasonably located, and provided with other convenience services. Second, the safety of the infrastructures should be considered, such as fire~\cite{clarke2019exploring} and flood~\cite{zaman2012regional,marvi2020review} risk, which periodically strike Southeast Australia. Third, the recommendations should consider the geographical and administrative characteristics of each local area to adapt to the local conditions and demand.

To construct a comprehensive view of EV charging station recommendations, we conduct a case study on NSW, Australia, and propose a data-driven, multi-source fusion system. To correspond to the requirements of well-accessible recommendations, the system refers to two-month EV trajectories in NSW and integrates data from multiple sources for a thorough evaluation of the recommended charging stations. The system can ensure the charging stations are located on routes/POIs, indicate the potential regional fire and flood risks, and perform separate analyses for each LGA to help with local government policy-making. Our system can be easily generalized to other regions or countries, allowing it to adapt to various climate and administrative conditions.

\vspace{-7pt}
\section{Related Work}
Various methods are developed to address the EV charging station location problem. Some divide the urban area into small regions to intuitively evaluate the charging demand for each local area. For example, Yi \textit{et al.} ~\cite{yi2020research} analyze the charging demand according to different urban functional areas. Ahn \textit{et al.} \cite{ahn2015analytical} divide the cities into square regions to derive the charging demand density. Others use simulation methods to model the growth of EV charging demand under current low EV penetration. Previous works use genetic algorithms~\cite{huang2020electric, efthymiou2017electric,jordan2022electric,zhu2016charging,zhou2022location} and whale optimization algorithm~\cite{cheng2022locating} to determine the amounts and locations of charging stations by simulating the charging demand distribution.

To enable evidence-based decision-making and uncover underlying behavior patterns of EV drivers, many studies utilize EV trajectory data to provide data-driven recommendations for optimal charging station locations.  
He \textit{et al.}~\cite{hehuan2021location} leverages GPS trajectory data to identify optimal electric taxi charging station locations by selecting the minimal number of facilities on the road network to cover all demand points, enhancing operational efficiency and sustainability. Bai \textit{et al.}~\cite{bai2019bi} strategically determines EV charging stations, capacities, and service types, ensuring comprehensive coverage of all potential charging demands within a grid-based urban framework. 
Liu \textit{et al.}~\cite{liu2019data} uses taxi trip data to explore travel patterns in a region and analyze the carbon dioxide emissions and residual electricity constraints associated with round-trips to charging stations.
Keawthong \textit{et al.}~\cite{keawthong2022location} utilizes queuing mode and travel time summarized from GPS data, pinpointing optimal charging stations and charger counts, and validates against real-world charging service data for efficiency.

Compared to existing approaches, our data-driven system integrates multi-source geospatial knowledge, allowing a holistic analysis of historical EV trips, ensuring environmental resilience, and considering complex urban planning requirements.

\section{System}

\subsection{Overview}

\begin{figure}[!h]

  \centering
  \includegraphics[width=0.95\linewidth]{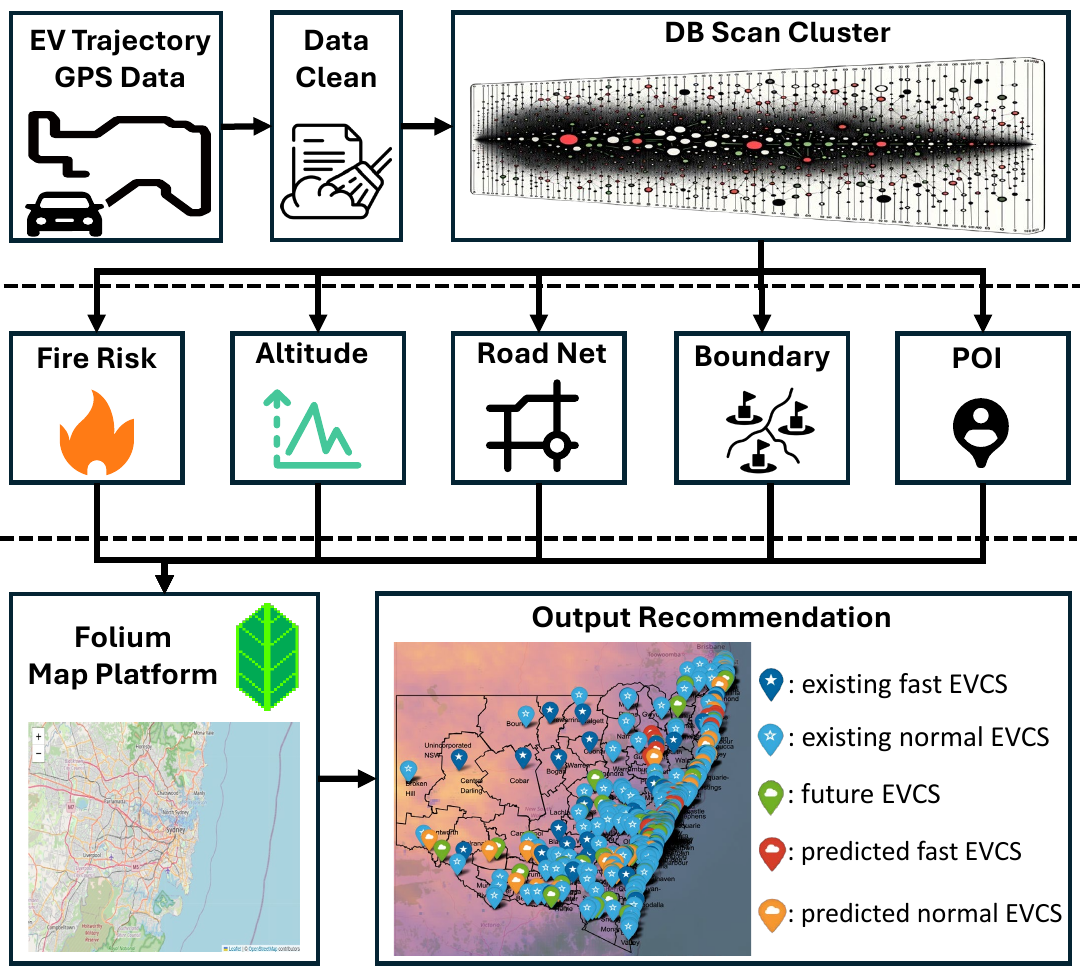}
  \vspace{-10pt}  % 适当调整这个值
  %最大0.89 
  \caption{The overview of our data-driven multi-source fusion system for EV charging stations (EVCS) recommendation. We utilize the NSW EV trips with data cleaning and apply the DBScan algorithm to recommend charging station locations according to EV trip points. Additionally, we overlay the fire risk map and display the nearest road segment altitude information for each existing and recommended charging station. We also do a separate analysis for each LGA and align our recommendations to the nearest POIs.}
  \vspace{-8pt} 
  \label{fig2}
\end{figure}

As shown in Figure~\ref{fig2}, we design a system that combines EV trip data and other various rich datasets and uses the advanced DBScan~\cite{ester1996density} algorithm to visualize our recommendations on Folium\footnote{\url{https://pypi.org/project/folium/}} platform.
\vspace{-5pt} 
\subsection{Multi-source Data Fusion}

% \subsubsection{Data description}
We will provide a detailed description of the data used in this section, as follows.

\begin{figure}[!ht]
    \centering
    
    \includegraphics[width=0.66\linewidth]{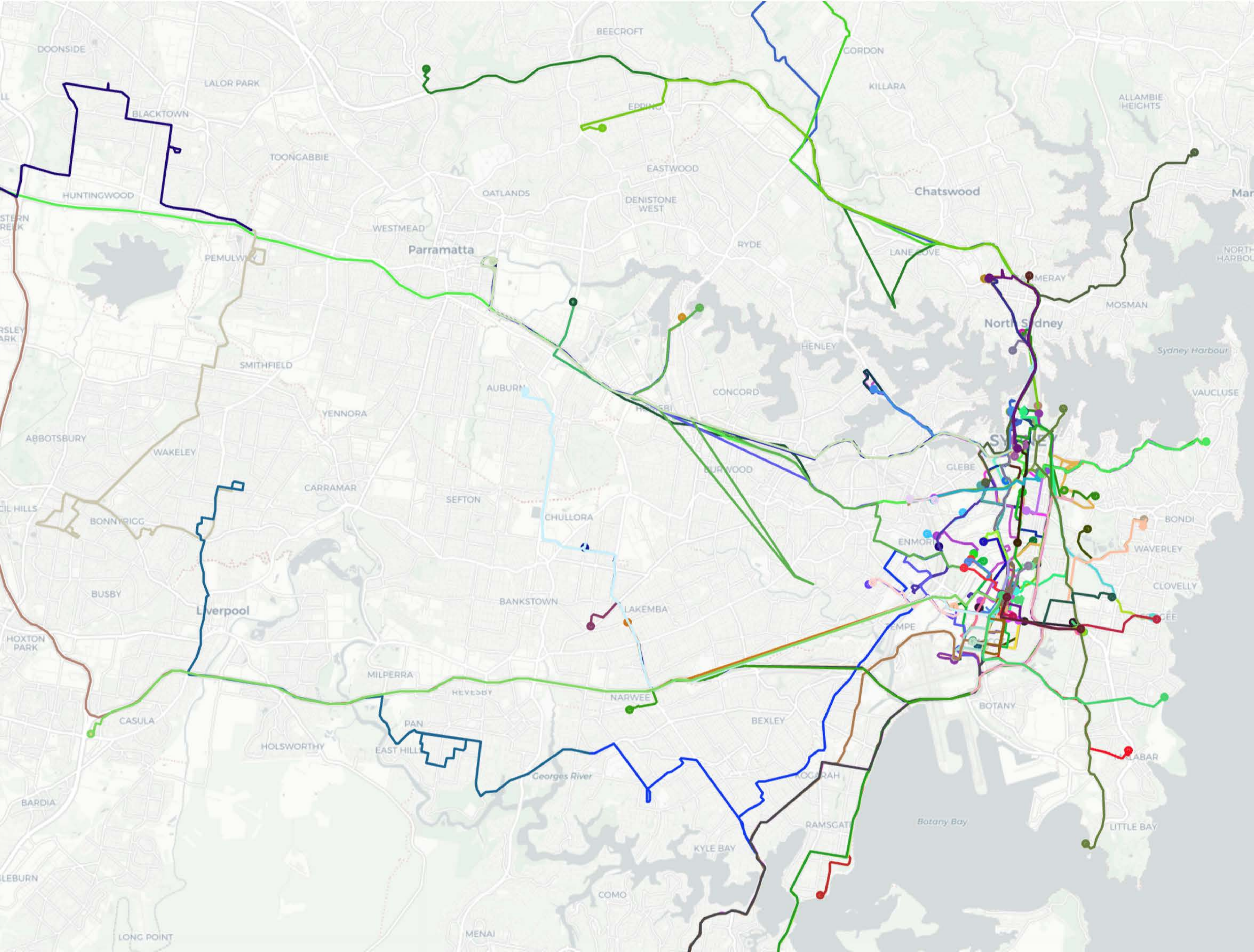}

% \subfloat[Statistics of EV trip data]{
% \includegraphics[width=0.48\linewidth]{fig/fig32.pdf}
% }
    % \vspace{-8pt} 
    \caption{The overview of EV trip data.}
    \setlength{\abovecaptionskip}{-12 cm} 
    \setlength{\belowcaptionskip}{-12 pt} 
    \label{fig3}
\end{figure}
\vspace{-2pt} 
% \textcolor{red}{visualization of the statistics of EV trip data.}

\noindent\textbf{NSW EV GPS trajectories.}
As shown in Fig~\ref{fig3}, NSW EV GPS trajectories data contains information on the driving trajectories of electric vehicles. Each record represents the detailed trip path of a single driving journey.

\begin{figure}[!ht]
\centering
\subfloat[EV charging stations]{
\includegraphics[width=0.47\linewidth]{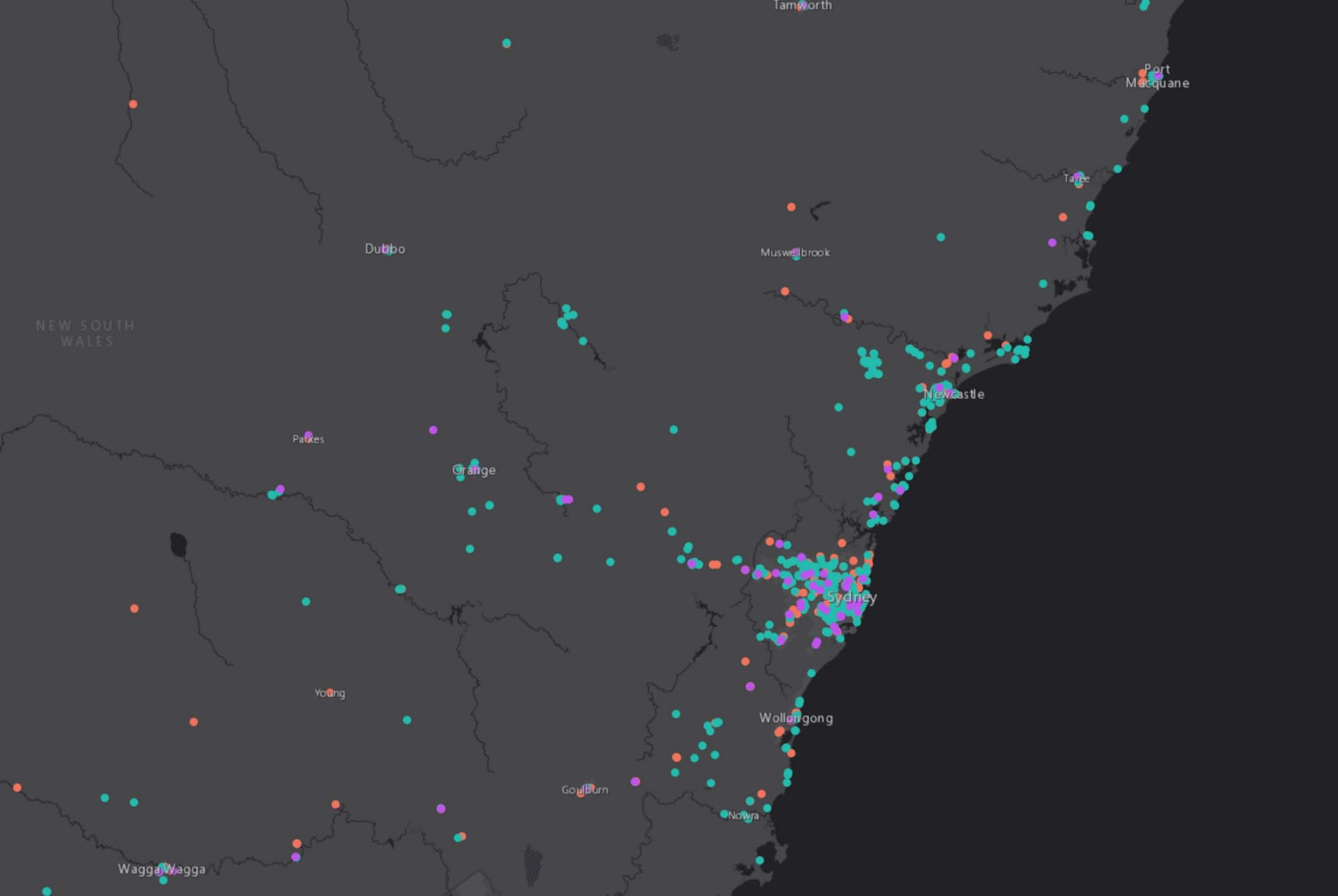}
\label{fig41}
}
\hfill
\subfloat[LGA boundary and route]{
\includegraphics[width=0.47\linewidth]{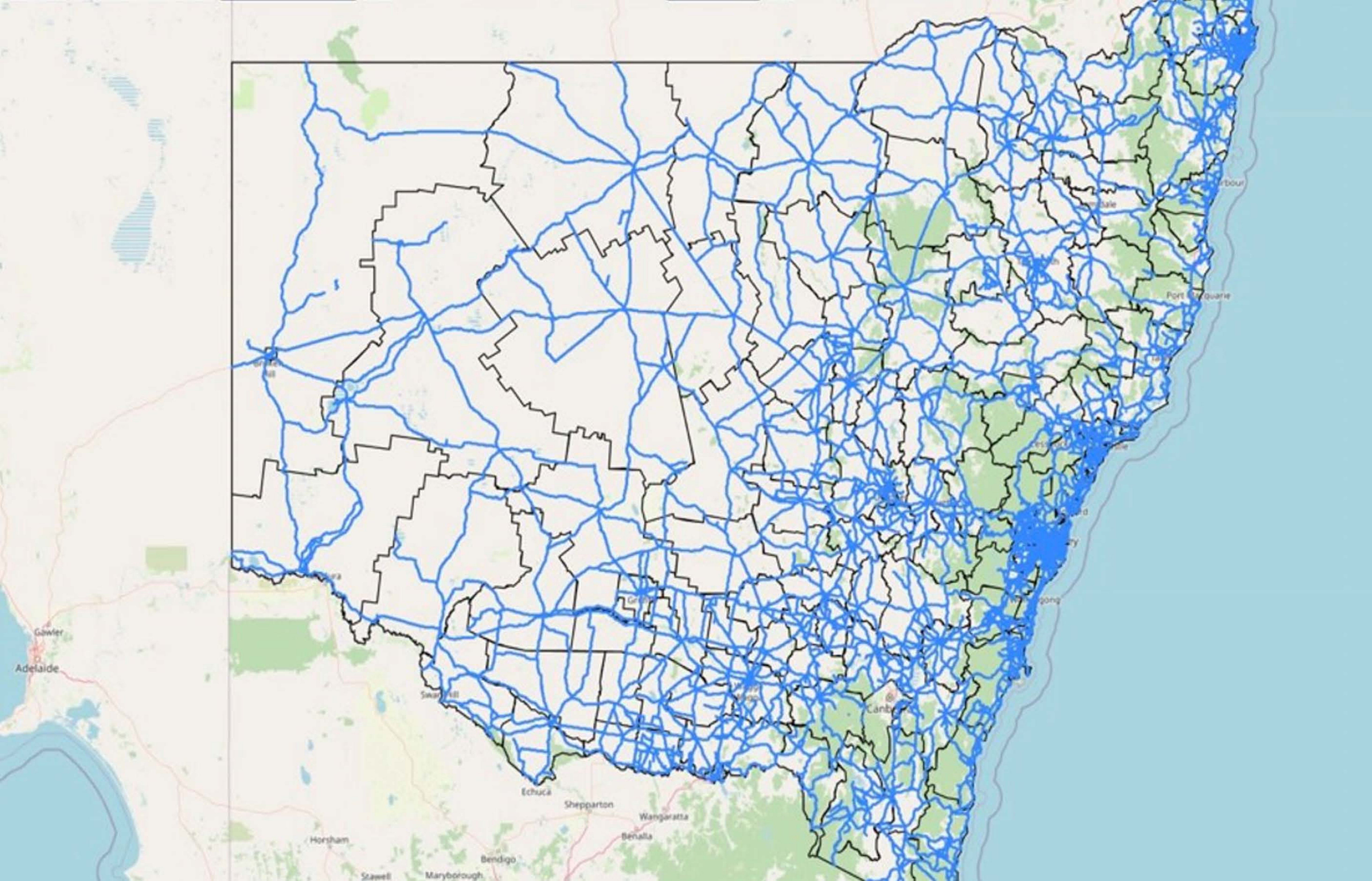}
\label{fig42}
}
\caption{Existing and approved EV stations, LGA boundaries, route map and altitude.}
\end{figure}

\noindent\textbf{Existing and approved EV stations.} We use the open-source electric vehicle station data made available by the NSW Government, as detailed in the NSW Master Plan\footnote{\label{master}\url{https://nswmaps.evenergi.com/}}, shown in as shown in Figure ~\ref{fig41},. This includes information on both destination and fast chargers in NSW, Australia, provided by the NSW Government. We use the existing charging stations for evaluations and comparisons with our recommended charging stations.

\noindent\textbf{LGA boundaries.}
LGA boundaries, as shown in Figure ~\ref{fig42}, released by the Australian Bureau of Statistics\footnote{\label{mynote}\url{https://www.abs.gov.au/}}, refer to the borders of local government areas in NSW. These boundaries are crucial for more informed planning and geographical analysis, contributing to charging demand allocation and policy-making for each local government.

\noindent\textbf{Route and altitude.}
Altitude information can be obtained from route data in the NSW Master Plan\footref{master}, and through preprocessing, altitude data is estimated to all points using the nearest neighbor approach. We assume the altitude information can reflect the potential flood risk.

\begin{figure}[!ht]
\centering
\subfloat[Climate (fire risk)]{
\includegraphics[width=0.47\linewidth]{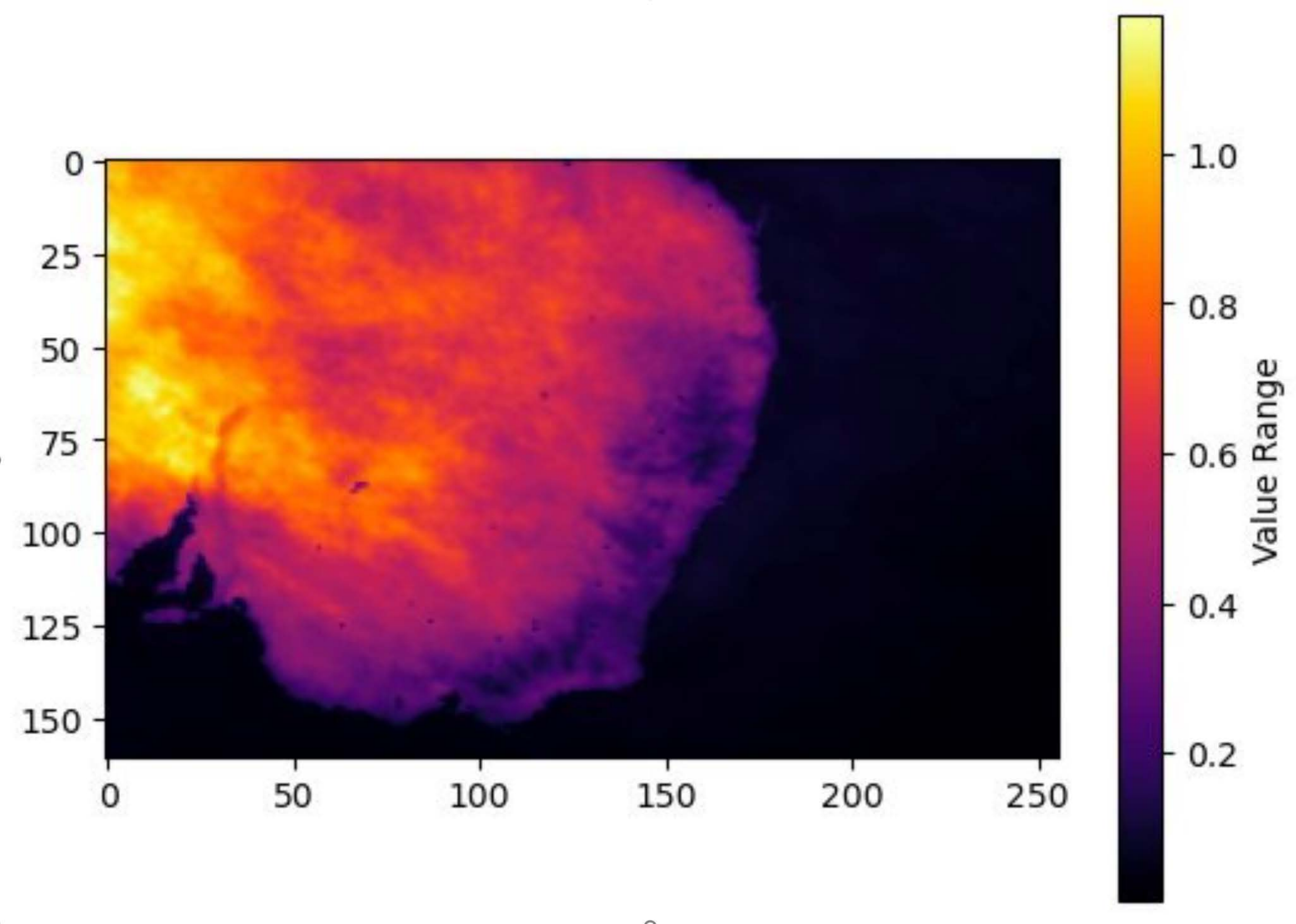}
\label{fig51}
}
\hfill
\subfloat[POIs]{
\includegraphics[width=0.47\linewidth]{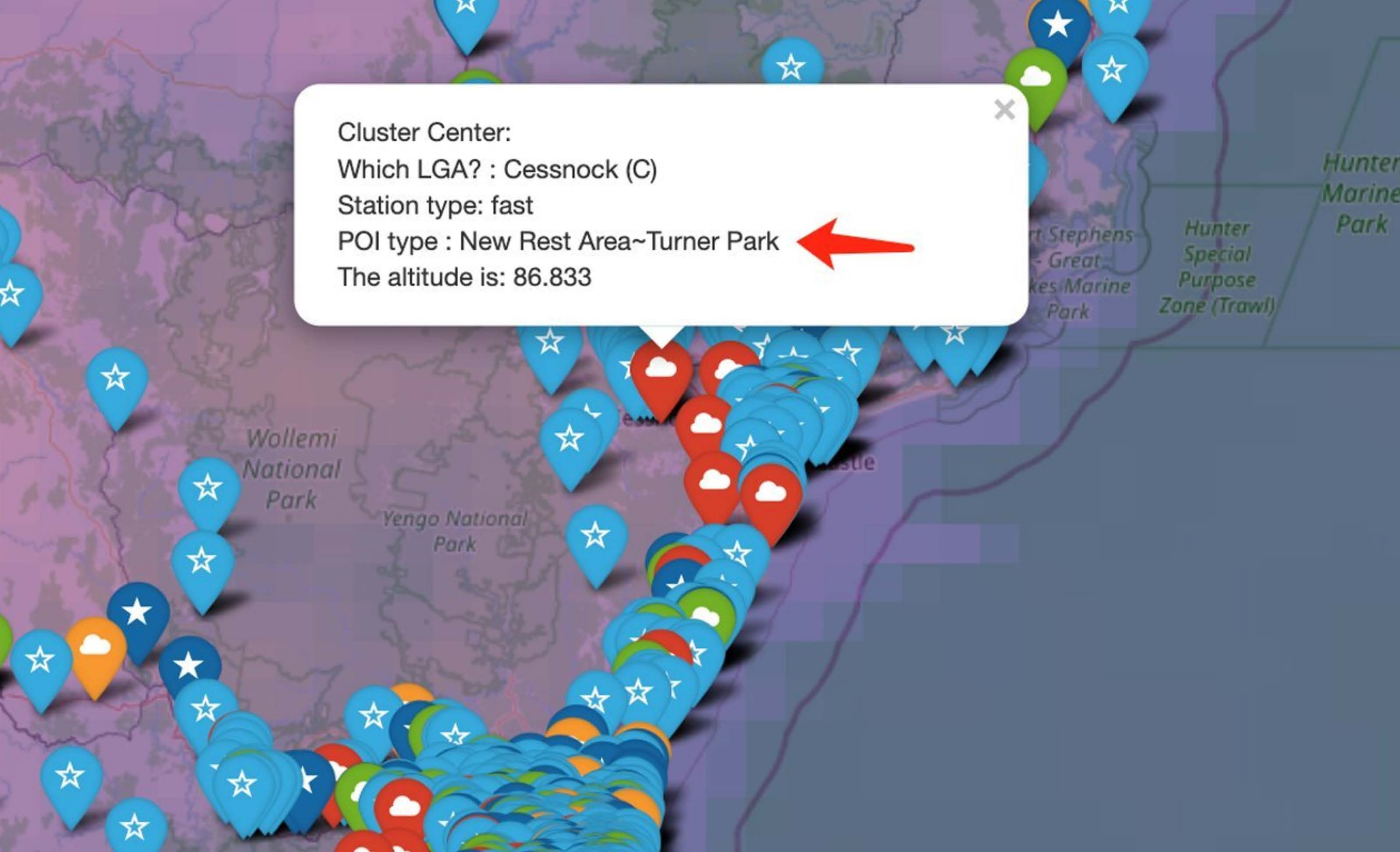}
\label{fig52}
}
\caption{Climate change projections map and POIs.}
\setlength{\abovecaptionskip}{-5 cm} 
\setlength{\belowcaptionskip}{-8 cm} 
\vspace{-10pt}
\end{figure}

\noindent\textbf{Climate.}
Climate (fire risk) data is sourced from the interactive climate change projections map of the Australian government website\footnote{\label{mynote}\url{https://www.climatechange.environment.nsw.gov.au/projections-map/}}, and as shown in Fig.~\ref{fig51}, the brightness indicates the heatmap showing the change of annual FFDI (forest fire danger index) from 2020 to 2039, where brighter areas are in higher risks of becoming fire-prone.

\noindent\textbf{POI.}
POI is also provided by NSW Master Plan\footref{master}, shown in Figure~\ref{fig52}. POIs contain fast food, fuel stations and tourism to provide references for aligning the recommended charging station locations considering the accessibility and utility.
\vspace{-12pt} 
\vspace{-10pt} 
\subsection{System Implementation} 
Due to the complexity of our data, including the heterogeneous distribution of the EV GPS trip data and the irregular shapes of the LGA boundaries, we employed the density-based DBScan algorithm for clustering.

Algorithm \ref{alg:dbscan} includes a set of data and parameters required such as LGAs, EV trips, searching radius $\epsilon$, the minimum number of points $MinPts$, and multi-source constraints mentioned in the previous sub-section. First, we iterate over the dataset within the boundaries of each LGA (line $1 \sim 2$). Building upon the traditional DBScan algorithm, it introduces an additional step to dynamically adjust the search radius $\varepsilon$ and the minimum number of points $MinPts$ (line 9) based on the altitude, POIs, and route information associated with each point. At last, the algorithm determines the location of the new cluster based on the previously estimated parameters (line 14).

\begin{algorithm}
\setstretch{1.0}
\caption{Enhanced DBScan Clustering Algorithm with Constraints for LGAs}
\label{alg:dbscan}
\begin{algorithmic}[1]
\Require Set of LGAs $L$, Dataset $D$, radius $\varepsilon$, minimum number of points $MinPts$, constraints $Constraints$
\Ensure Cluster indices for each LGA in dataset $D$
\For{each $lga \in L$}
    \State Extract $D_{lga}$ from $D$ within the boundaries of $lga$
    \State Initialize $C = 0$ \Comment{Cluster counter}
    \State Initialize $visited = \emptyset$, $noise = \emptyset$
    \For{each point $p \in D_{lga}$}
        \If{$p$ is not in $visited$}
            \State Add $p$ to $visited$
            \State $NeighborPts = \{q \in D_{lga} | \text{distance}(p, q) \leq \varepsilon\}$
            \State Apply constraints $Constraints$ to adjust $\varepsilon$ and $MinPts$ based on $p$'s altitude, POI, and route information
            \If{sizeof($NeighborPts$) $< MinPts$}
                \State Add $p$ to $noise$
            \Else
                \State $C = C + 1$
                \State ExpandCluster($p$, $NeighborPts$, $C$, $\varepsilon$, $MinPts$)
            \EndIf
        \EndIf
    \EndFor
\EndFor
\end{algorithmic}
\end{algorithm}
% \DecMargin{0 em}
\setlength{\textfloatsep}{0.1cm}
\setlength{\floatsep}{0.2 cm}

\vspace{-8pt} 

\begin{figure}[!ht]
    \centering
    \includegraphics[width=0.87\linewidth]{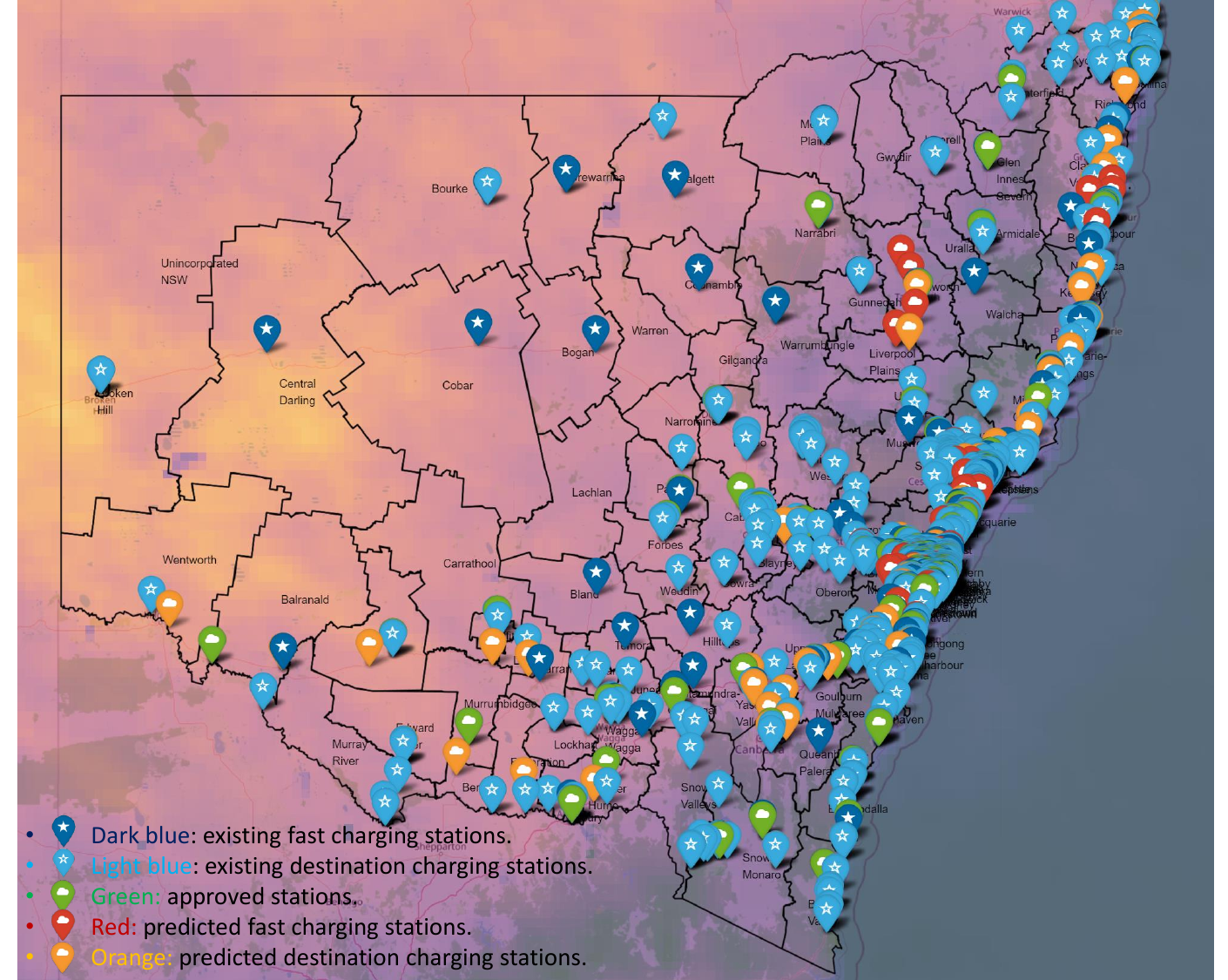} % Adjust the width as necessary
    \caption{Visualization of all EV charging stations in NSW.}
   
    \setlength{\belowcaptionskip}{-1.2 cm} 
    \label{fig:NSW}
\end{figure}

\begin{figure}[!ht]
    \centering
    \includegraphics[width=0.83\linewidth]{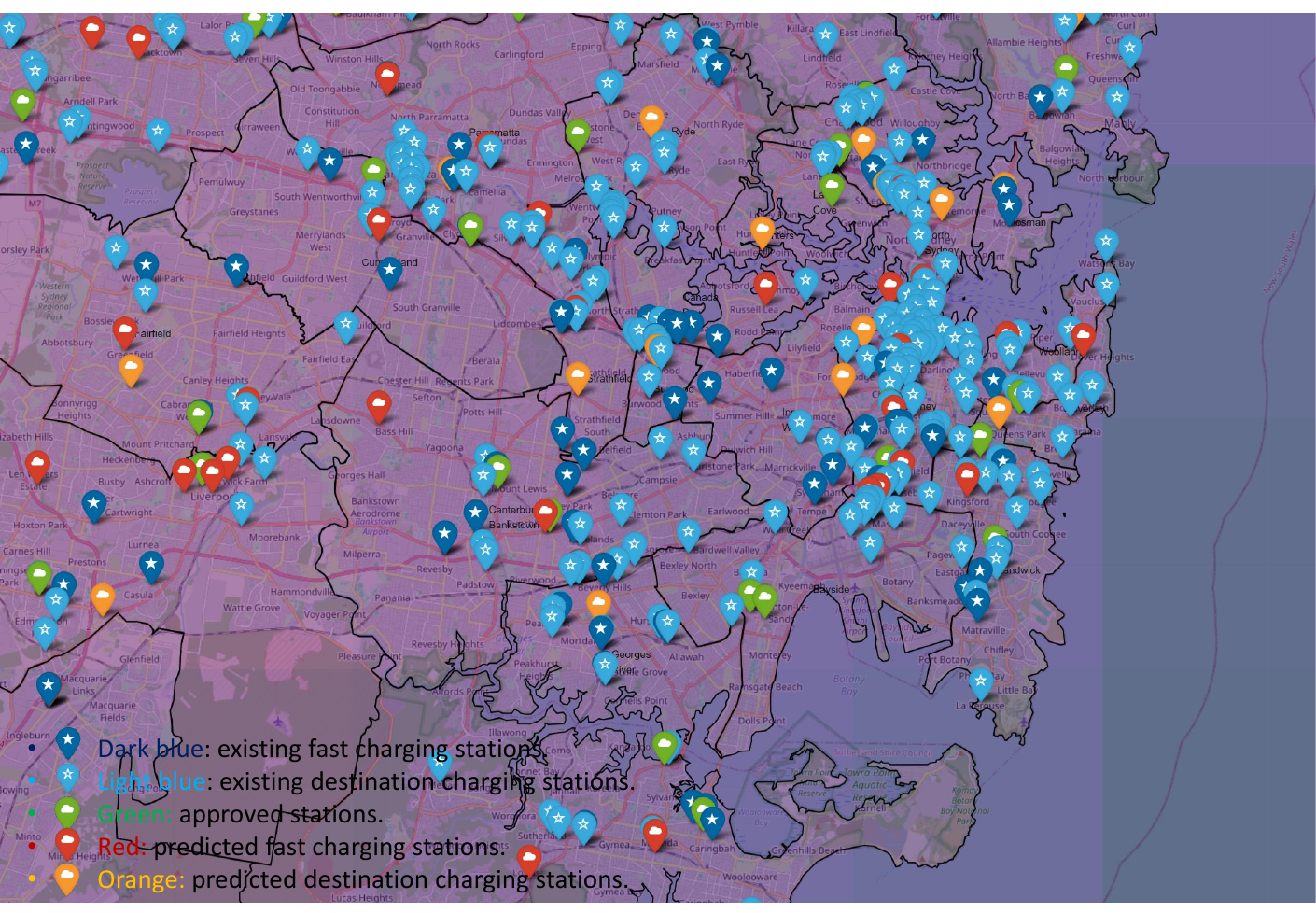} % Adjust the width as necessary
    \caption{Visualization of EV charging stations in the urban areas of NSW.}
   \setlength{\abovecaptionskip}{-1. cm} 
    \setlength{\belowcaptionskip}{-1. cm} 
    \label{fig:urban}
\end{figure}
% \vspace{-3pt} 

\section{Experimental Analysis}

\subsection{Visualization of EV charging stations recommendations}
We visualize the recommended locations with an interactive map rendering tool ``folium''. Fig. \ref{fig:NSW} is the visualization of both existing and recommended charging stations in NSW. In addition to the existing 263 fast (dark blue markers) and 785 destination (light blue markers) charging stations in the NSW master plan, we have successfully recommended 93 fast charging (red markers) and 103 destination (orange markers) charging stations. The green markers represent the currently approved charging stations and the black lines are the LGA boundaries. 

When clicking on the markers, they show the basic information shown in Fig. \ref{fig:case1} about this station, including the altitude. We can see that both existing and recommended stations are distributed more on coastal than inland LGAs. This is due to the much larger number of developed cities near the East Coast having more demand. The FFDI map is overlaid to give a straightforward review of the fire risk of each station. Since the charging stations are too dense in Sydney City, we provide a zoom-in view shown in Fig. \ref{fig:urban}. We can find that all the recommended stations are located on road segments, which benefits from the POI constraints as they are usually located near the roads.

\vspace{-8pt} 

\subsection{Case Study}
\begin{figure}[!ht]
\centering
\subfloat[Case study 1: Ours vs. the existing NSW Master Plan.]{
\includegraphics[width=0.46\linewidth]{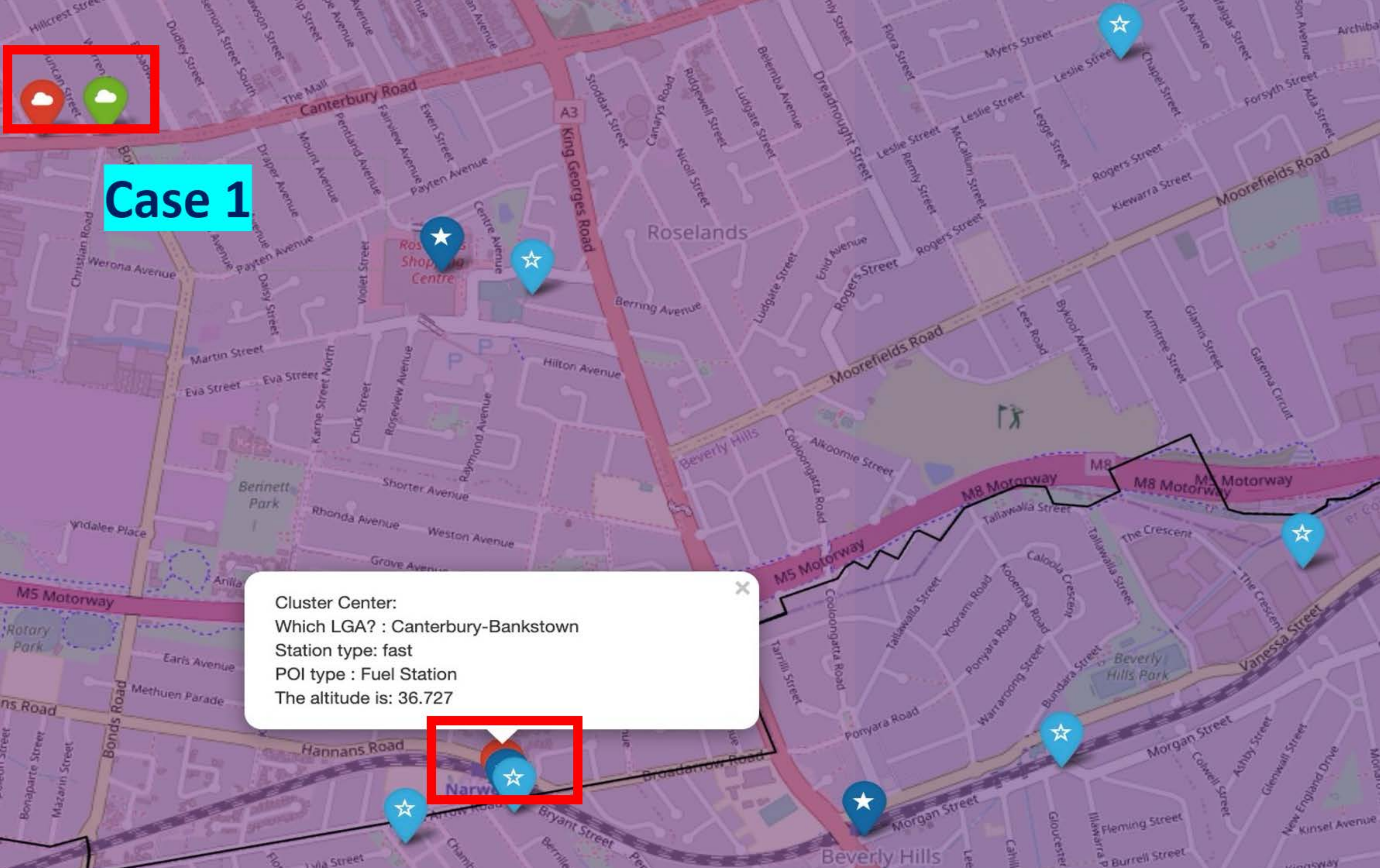}
\label{fig:case1}
}
\hfill
\subfloat[Case study 2: Ours vs. the existing NSW Master Plan.]{
\includegraphics[width=0.46\linewidth]{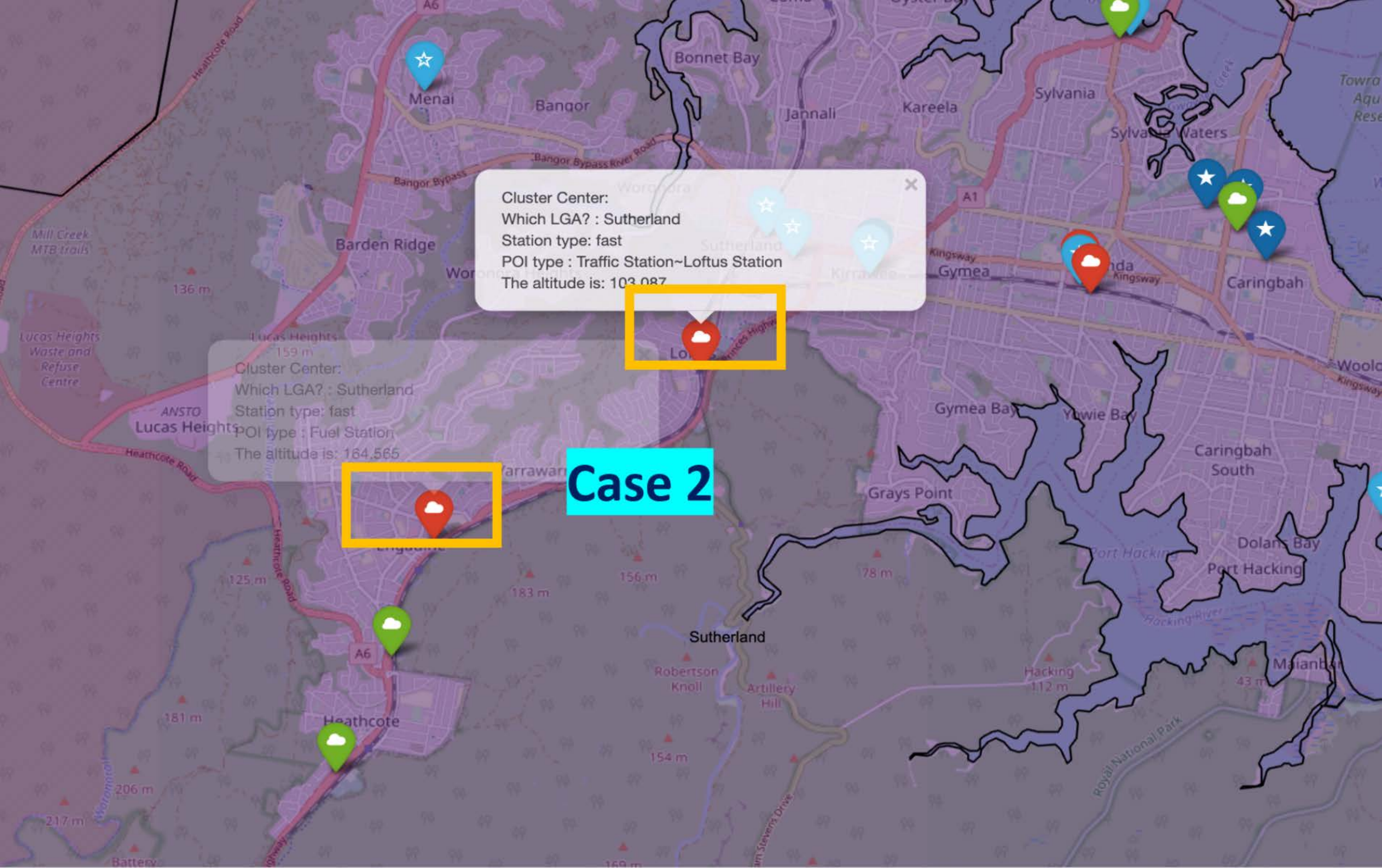}
\label{fig:case2}
}
\caption{Case Study.}
\setlength{\belowcaptionskip}{-8 cm} 
\end{figure}

We use two cases to show some improvement compared to the master plan. Case 1 in Fig. \ref{fig:case1} demonstrates our recommendations can align with the existing and approved charging stations, showing that our system can recommend reasonable stations from EV trip data. Case 2 illustrates two charging stations recommended in the area with no existing or approved stations. These two recommendations are located in the centers of two suburbs due to their business attributes, proving that we can infer more new charging stations that reflect the local demand by capturing both EV trip locations and density in each LGA. As our clustering-based model is data-driven, the lack of inland historical EV trip data limits the recommendations on those LGAs, resulting in sparse recommendations in western inland LGAs. Therefore, we can further enhance our recommendations by modeling with more supplementary data (e.g., population density) that can reflect the future demand of those rural areas.

\vspace{-20pt} 
\section{Conclusion}
In conclusion, our case study in NSW, Australia, demonstrates the efficacy of a data-driven, multi-source fusion system in strategically placing EV charging stations. By integrating EV trajectory data with geographical, environmental, and administrative factors, our approach ensures that charging stations are accessible, safe, and optimally located. This comprehensive methodology not only supports local government policy-making but also provides a scalable framework that can be adapted to different regions and countries, considering their unique climatic and administrative conditions. As the adoption of EVs continues to rise, such innovative systems will be essential in meeting the growing charging demands and fostering a sustainable future.

For future work, we may improve the system if we are authorized with more detailed data, including the electricity supply~\cite{ahmad2022optimal} near the Sydney area to consider charging capacity, the existing charger types and urban functional area divisions~\cite{yi2020research} to classify fast/destination charging stations, and so on. These data can help with classifying the recommended charger types and evaluating the maximum acceptable load of charging stations.
\vspace{-5pt}
\begin{acks}
We would like to acknowledge the support of Cisco's National Industry Innovation Network (NIIN) Research Chair Program, the support from Transport for NSW, and Compass IoT.
\end{acks}
\vspace{-5pt}
\bibliographystyle{ACM-Reference-Format}
\bibliography{sample-base}


\end{document}